\documentclass[letterpaper, 10 pt, conference]{ieeeconf}  

\IEEEoverridecommandlockouts                              

\usepackage{graphics} 
\usepackage{epsfig} 
\usepackage{times} 
\usepackage{amsmath} 
\usepackage{amssymb}  
\usepackage{booktabs}
\usepackage{todonotes}
\usepackage{siunitx}
\usepackage{array,multirow}
\usepackage{makecell}

\makeatletter
\let\NAT@parse\undefined
\makeatother
\usepackage{hyperref}  
\usepackage{cleveref}

\title{\LARGE \bf
SKooP: Symmetric Koopman Predictions for Faster and More Generalizable Legged Robot Locomotion with Reinforcement Learning
}

\author{Evelyn D'Elia$^{1,2}$, Weishu Zhan$^{2}$, Giulio Turrisi$^{3}$, Giulio Romualdi$^{4}$, Giuseppe L'Erario$^{4}$, \\ Raffaello Camoriano$^{5,6}$, Wei Pan$^{7}$, and Daniele Pucci$^{4}$
\thanks{*This study was carried out within the FAIR - Future Artificial Intelligence Research and received funding from the European Union Next-GenerationEU (PIANO NAZIONALE DI RIPRESA E RESILIENZA (PNRR) – MISSIONE 4 COMPONENTE 2, INVESTIMENTO 1.3 – D.D. 1555 11/10/2022, PE00000013). This manuscript reflects only the authors’ views and opinions, neither the European Union nor the European Commission can be considered responsible for them. }
\thanks{$^{1}$IIT@MIT, Italian Institute of Technology (IIT), 16163 Genoa, Italy
        {\tt\small evelyn.delia@iit.it}}%
\thanks{$^{2}$Machine Learning and Optimisation, University of Manchester, M13 9PL Manchester, U.K.}%
\thanks{$^{3}$Dynamic Legged Systems Lab, IIT, 16163 Genoa, Italy}
\thanks{$^{4}$Generative Bionics S.R.L, 16163 Genoa, Italy}
\thanks{$^{5}$DAUIN, Politecnico di Torino, 10129 Turin, Italy}%
\thanks{$^{6}$Rehab Technologies Lab, IIT, 16163 Genoa, Italy}%
\thanks{$^{7}$School of Engineering, Newcastle University, NE1 7RU Newcastle upon Tyne, U.K.}%
\thanks{E. D'Elia, G. Romualdi, G. L'Erario, and D. Pucci contributed to this work while at the Artificial and Mechanical Intelligence lab, IIT, Italy.}%
\thanks{W. Pan contributed to this work while at the Machine Learning and Optimisation group, University of Manchester, U.K.}%
}

\begin{document}

\maketitle
\thispagestyle{empty}
\pagestyle{empty}

\begin{abstract}


Reinforcement learning (RL) algorithms classically suffer from poor sample efficiency.
In robotics, a recent line of work has emerged addressing this problem by encoding physics priors in the learning process.
However, most of these approaches are validated on well-defined, low-dimensional benchmark systems rather than high-dimensional robots with complex nonlinear dynamics.
In this paper, we introduce \textit{SKooP (Symmetric Koopman Predictions)}, an approach combining the advantages of morphological symmetries with those of a Koopman model learned via autoencoder to enhance policy learning.
SKooP learns a Koopman model of the system dynamics alongside the policy. 
The resulting Koopman predictions are used as privileged observations for the critic, allowing the agent to learn based on smoother, more informative features.
We also incorporate group symmetries into the actor, critic, encoder and decoder networks to produce a highly equivariant policy.
The SKooP approach is validated via in-depth analysis of the learned Koopman models and symmetric policies to showcase how each of these influences the agent's performance. We also show that the learned policies are transferable to different simulation environments.
Our results show that SKooP consistently reduces convergence time and increases the learned reward for multiple challenging bipedal locomotion tasks on 
a quadruped robot.
Project page: \url{https://evelyd.github.io/SymmetricKoopmanPredictions}

\end{abstract}

\section{INTRODUCTION}

The use of reinforcement learning approaches for robotic control has skyrocketed in recent years, mainly thanks to advances in computational efficiency and simulation fidelity. 
However, for systems such as legged robots which have complex, nonlinear dynamics, learning effective policies is still a challenging and active research area.
Although powerful, purely data-driven model-free RL approaches rely on costly trial and error. 
Conversely, more traditional model-based control approaches exploit the known physics of the system, but typically require expert knowledge and extensive manual tuning.
Fusing model-based control with RL holds promise for overcoming their respective limitations and devising more efficient and effective robot control methods.

\emph{Model-based} control methods employ a complete or reduced description of robot dynamics. 
Using the complete dynamics model requires explicit design and precise modeling, but produces a high-fidelity result. 
Instead, reduced or simplified model-based approaches, e.g., with inverted pendulum~\cite{Kajita} or centroidal dynamics~\cite{Englsberger2015, Dai2014} models, improve computational complexity while sacrificing fidelity.

\emph{Reinforcement learning}, conversely, can be \emph{model-free} thanks to the parallelized, high-fidelity simulators that are nowadays available, and recent works show its success in learning legged robot tasks \cite{Radosavovic2024, Hwangbo2019}. Nonetheless, model-free strategies suffer from low sample efficiency, lack of generalization, and convergence to suboptimal policies.
In recent years, injecting physics priors into the RL training process has emerged as a strategy to mitigate the drawbacks of the two separate approaches.
Here, we utilize two specific types of priors: symmetry information and linearized dynamics in the Koopman-lifted space~\cite{Koopman1931}.

\begin{figure}
\vspace{0.2cm}
    \centering
    \includegraphics[width=\linewidth]{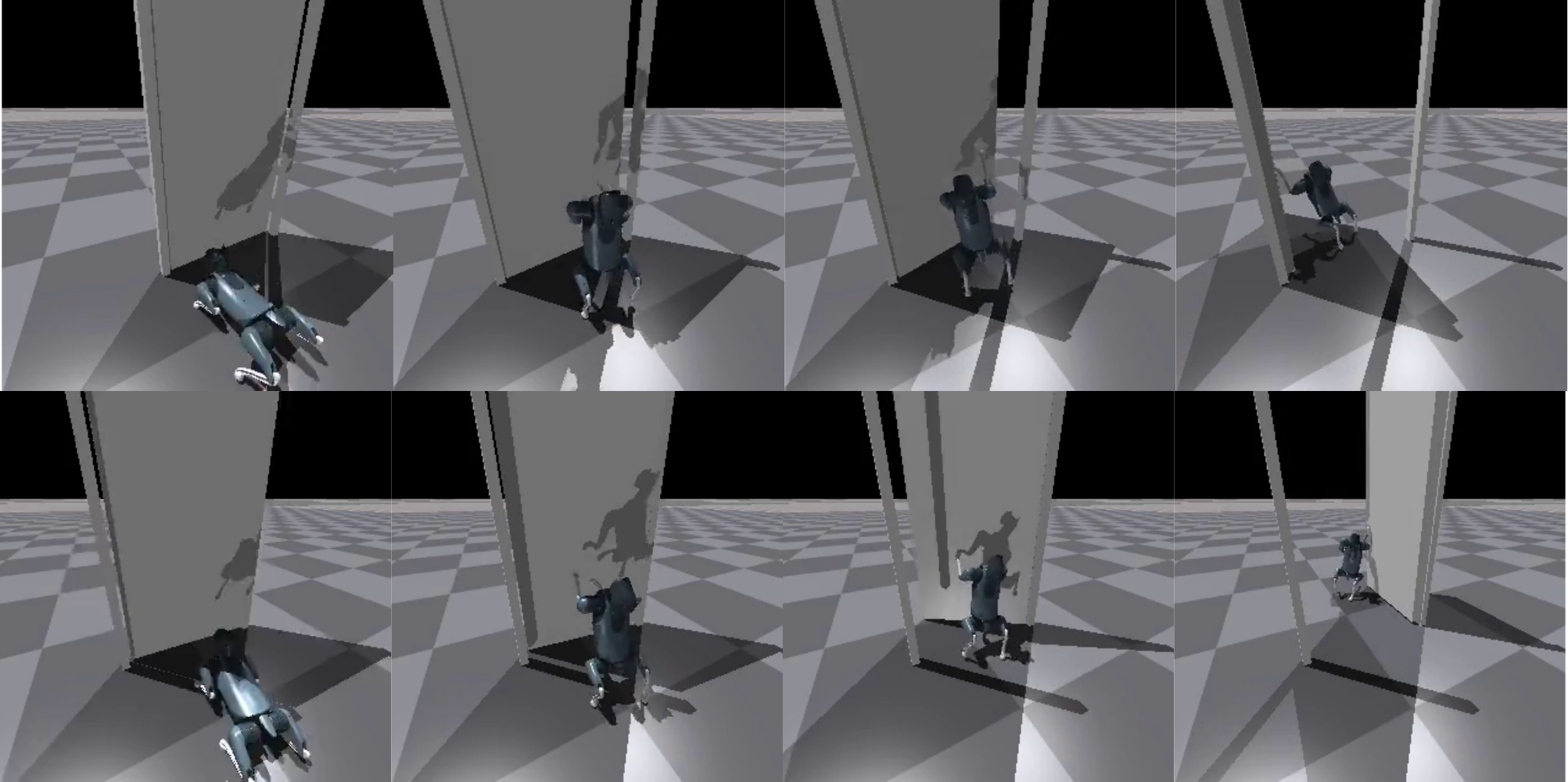}
    \vspace{-0.5cm}
    \caption{Comparison of SKooP performance on trained vs. mirrored \emph{push door} task. Top row: right-opening door (training task). Bottom row: left-opening door (unseen symmetric task).}
    \label{fig:lr_push_door}
    \vspace{-0.5cm}
\end{figure}


The use of symmetries as priors was first introduced for supervised deep learning, 
namely through the imposition of symmetry constraints on learning architectures~\cite{Weiler2023, Gerken2023, deHaan2024}.
In robot learning, a handful of works incorporate symmetries into the RL process via data augmentation~\cite{Mittal2024, Su2024}.
The RL framework proposed in~\cite{Su2024} constrains the learned policy to be equivariant. 
Encoding symmetry information in this way reduces the number of samples required for training and yields more generalizable policies, but introducing such constraints can negatively affect performance.

The other type of prior we consider is based on Koopman theory~\cite{Koopman1931}, which states that for any nonlinear system there exists an infinite-dimensional function space in which the system dynamics are globally, not just locally, linear. 
Early data-driven methods found finite approximations of the Koopman operator, such as via dynamic mode decomposition (DMD)~\cite{Tu2014} and DMD with control (DMDc), which extends the theory to controlled systems to produce a controlled Koopman model~\cite{Proctor2018}.
More recent works approximate both the lifting functions and the linear embedding with kernel methods or autoencoders~\cite{Williams2015, Lusch2018}.
It has been shown that the controlled Koopman system can be effectively regulated with linear model predictive control (MPC)~\cite{Korda2018, Korda2020}.

Some recent works also utilize Koopman priors for RL. For instance,~\cite{Song2021, Mayfrank2025} use the Koopman operator to facilitate learning optimal controllers. \cite{Rozwood2023} presents two strategies in which the value function constitutes a linear mapping of the Koopman observables, and shows these methods to have state-of-the-art policy-learning performance on benchmark systems such as the Lorenz model~\cite{lorenz1963deterministic}.
Another approach, proposed in~\cite{Cozma2025}, trains the Koopman autoencoder in parallel with Proximal Policy Optimization (PPO)
and learns the policy directly from 
the encoded Koopman state, achieving modest performance improvements on RL benchmark tasks. Although the results are relevant from a theoretical perspective, they are validated only on well-defined and low-dimensional systems.

In robotics, high-dimensional state spaces can be lifted and linearized at relatively low cost with an autoencoder, bypassing the hand-design of lifting functions. 
Many of the Koopman-based approaches for robotics specifically focus on characterizing the dynamics~\cite{Apraez2024} or tailor the linearization of the system for use in optimal control frameworks such as MPC~\cite{Li2024}. Some approaches use the Koopman operator with RL to learn the optimal control tuning parameters rather than the actions directly, e.g., \cite{Martini2025}.
However, none of these robotic control approaches directly inject the Koopman prediction into the policy learning process.

A key motivation for our work is that \emph{Koopman theory can facilitate the identification and exploitation of symmetries.} 
For example,~\cite{Weissenbacher2022} improves the sample efficiency of an offline RL algorithm by using the Koopman linearization to more effectively enforce symmetries in the system via data augmentation, 
validating the approach on RL benchmark tasks.
In~\cite{Apraez2024}, a system's morphological symmetries are predefined and embedded in an equivariant Koopman autoencoder to more efficiently learn a global linear model. The authors present a harmonic analysis of quadruped locomotion modes, but do not address the control of the robot.

In this work, we propose \textbf{SKooP} (\textbf{S}ymmetric \textbf{Koo}pman \textbf{P}redictions) an approach which combines the advantages of a controlled Koopman embedding and morphological symmetry priors to enable  faster initial policy convergence and improve learned behavior quality and generalization. By learning a symmetric linearized latent space concurrently with the policy, we ensure that the lifting function is trained on a relevant state space. Then, by providing the critic with information about the future in the form of Koopman-predicted latent states, we enable it to learn higher-quality motions.
Specifically, our contributions are as follows:
\begin{itemize}
    \item A \emph{control- and symmetry-informed Koopman autoencoder} to predict the future state, trained concurrently with the RL policy.
    \item A \emph{modular, symmetric Koopman-based RL architecture}, SKooP, which learns the Koopman embedding and employs the lifted state prediction as a prior for learning the value function.
    \item \emph{Detailed ablation analysis and performance validation} on multiple challenging two-legged locomotion tasks for a quadruped robot, in two different simulation environments.
\end{itemize}

Our results show that SKooP improves initial reward convergence rate and symmetry of learned motions compared with standard PPO, and that it generalizes better than a state-of-the-art symmetric RL method to unseen, mirrored tasks.

\section{BACKGROUND}
\subsection{Notation}
\begin{itemize}
    \item $g \in \mathbb G$ denotes a group element within a group representation. Specifically, $\mathbb C_2 := \{e, g_s | g_s^2 = e\}$ is a reflection group; $g_s$, $e$ are the reflection and identity transformations.
    \item $\rho_{\mathcal X}, \rho_{\mathcal U}, \rho_{\mathcal Y}$ are matrix representations of a symmetry group acting on the state space $\mathcal X \subset \mathbb{R}^{n'}$, the control space $\mathcal U \subset \mathbb{R}^m$, and the output space $\mathcal Y \subset \mathbb{R}^{n' \times n}$, respectively. 
    \item $\mathcal K$ is the Koopman operator, which acts on observables $z := \psi(x) := [\psi_1(x), \dots, \psi_n(x)]$, where $\psi_i(x) \in \mathcal F(\mathcal X)$, to predict the system dynamics $F(x_k, u_k)$ in the lifted space.
    \item $A \in \mathbb R^{n\times n}$, $B \in \mathbb R^{n\times m}$, $C \in \mathbb R^{n'\times n}$ are linear system matrices and $H \in \mathbb N$ is the look-ahead horizon length.
    \item $q, \dot q$ are joint positions and velocities.
    \item $\theta$, $\xi$, $\phi$ represent learned network parameters.
    \item $\mathcal L_{sr}$, $\mathcal L_{lp}$, $\mathcal L_{sp}$ are the standard autoencoder losses: state reconstruction, latent state prediction, and state prediction, respectively.
\end{itemize}

\subsection{Markov Decision Processes and Bellman Equation}

Markov Decision Processes (MDPs) are used in optimal control and RL for modeling transition-based decisions. 
An MDP is represented by the tuple $(\mathcal X, \mathcal U, \mathcal P, r, \gamma)$ where $\mathcal P : \mathcal X \times \mathcal U \times \mathcal X \mapsto [0, 1]$ is the transition probability, $r: \mathcal X \times \mathcal U \times \mathcal X \mapsto \mathbb R$
is the reward function, and $\gamma \in [0, 1]$ is the discount factor. RL aims to find an optimal policy $\pi^*$ that maximizes the Bellman equation:
\begin{equation}
    V_{\pi}(x) = \sum_{x' \in \mathcal X} \mathcal P(x'|x, \pi(x)) \left[r(x, \pi(x), x') + \gamma V_{\pi}(x')\right],
    \label{eq:value}
\end{equation}
where $V_{\pi}(x)$ is the value of state $x$ using policy $\pi$.

\subsection{Morphological Symmetries}
\label{subsec:background_symmetries}
In this work, we take advantage of the inherent symmetries of the state and action spaces of the robot. 
Most legged robots possess $\mathbb C_2$ symmetry over the sagittal plane, allowing information such as joint positions and poses to be mirrored over this plane.
A symmetry acting on a point $x \in \mathcal X$ can be interpreted as a matrix-vector multiplication with $g \triangleright x := \rho_{\mathcal X}(g) x \in \mathcal X$.

A map $f$ is considered \emph{group invariant}, or \emph{$\mathbb G$-invariant}, if its output does not change under any group transformation of the input, i.e., $
f(\rho_{\mathcal X}(g) x) = f(x), \; \forall g \in \mathbb G$, while it is \emph{$\mathbb G$-equivariant} if applying a group transformation to the input and then evaluating $f$ yields the same result as applying the corresponding group transformation to the output, i.e., $
f(\rho_{\mathcal X}(g) x) = \rho_{\mathcal Y}(g) f(x), \; \forall g \in \mathbb G$.

In the context of MDPs, the state and action group representations $\rho_{\mathcal X}$ and $\rho_{\mathcal U}$ can be defined as symmetric representations given the morphological symmetries of the system. This definition proves useful in the context of RL, where it makes the policy $\pi : \mathcal X \mapsto \mathcal U$ $\mathbb G$-equivariant and the reward function $r : \mathcal X \times \mathcal U \mapsto \mathbb R$ $\mathbb G$-invariant. 
Thus, the value function $V_{\pi} : \mathcal X \mapsto \mathbb R$ is also $\mathbb G$-invariant \cite{Ordonez2025}.

An unknown target function to approximate via a neural network can be expressed as $f_{\phi} \in \mathcal F : \mathcal X \mapsto \mathcal Y$. To impose symmetry constraints, we can require that the learned function $f_{\phi}$ be $\mathbb G$-invariant or $\mathbb G$-equivariant:
\begin{equation}
    \begin{aligned}
    \mathcal F_{\mathbb G}^{inv} &:= \{f_{\phi} \in \mathcal F | y = f_{\phi}(g \triangleright x), \forall g \in \mathbb G \}, \\
    \mathcal F_{\mathbb G}^{eq} &:= \{f_{\phi} \in \mathcal F | g \triangleright y = f_{\phi}(g \triangleright x), \forall g \in \mathbb G \}.
\end{aligned}
\end{equation}

\subsection{Koopman Operator}

The Koopman operator offers a linear representation of the
nonlinear system dynamics $x_{k+1} = F(x_k)$:
\begin{equation}
    \begin{aligned}
        \mathcal K \psi({x_k}) &:= \psi( F( x_k)) = \psi(x_{k+1}), \\
        z_k &:= \psi(x_k). 
    \end{aligned}  
\end{equation}
A key challenge in finding a Koopman model is to design the lifting function $\psi(x)$. 
Given that we must approximate an infinite-dimensional Koopman operator, the latent space $\mathcal F(\mathcal X)$ is much higher-dimensional than the state space $\mathcal X$ \cite{Brunton2022}.
Moreover, \cite{Korda2018} proposes an extension of the Koopman operator 
approximation
for controlled systems $x_{k+1} = F(x_k, u_k)$:
\vspace{-0.1cm}
\begin{equation}
    \begin{aligned}
    z_{k+1} &= A z_k + B u_k, \\
    x_k &= C z_k.
\end{aligned}
\label{eq:controlled_koopman}
\end{equation}
For a controlled system, the linear representation is supported on the policy control set.
As such, finding an approximation of the Koopman operator $\mathcal K$ amounts to finding $[A, B]$. 
Extending Koopman theory to the controlled case opens up new possibilities for its use in optimal control and RL to provide predictive priors. 

\section{METHODS}

\begin{figure}
\vspace{0.2cm}
    \centering
    \includegraphics[width=0.99\linewidth]{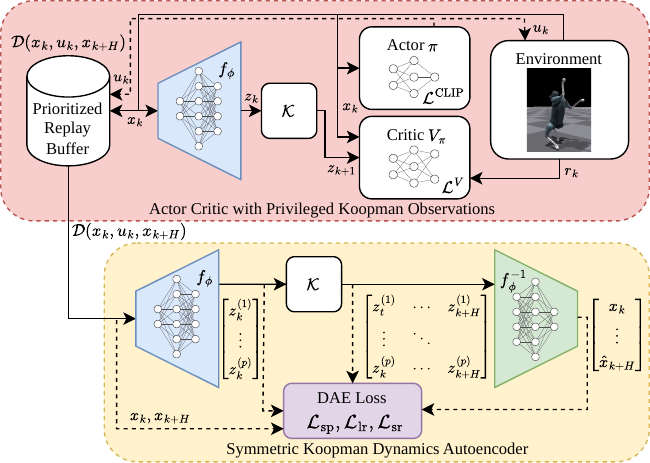}
    \vspace{-0.5cm}
    \caption{Schematic representation of SKooP.}
    \label{fig:diagram}
    \vspace{-0.5cm}
\end{figure}

In this Section, we present our novel, modular, and online SKooP approach that exploits symmetries and Koopman theory to enhance the sample efficiency and improve the performance of actor-critic RL algorithms.
Our approach utilizes a symmetric Koopman dynamics autoencoder to learn a lifted space in which the dynamics are linear and system symmetries are respected. The inclusion of a Koopman prediction as
part of the critic input in an actor-critic framework enables the critic to more easily learn a precise value function. 
This is thanks to the 
linear representation of the system dynamics that the Koopman model provides, which simplifies the critic's learning problem of predicting the expected future return.
The overall scheme of the proposed approach is displayed in~\Cref{fig:diagram}.

\subsection{Equivariant Networks}
Our approach implements symmetry priors in two ways. The first way is via the structure of the actor and critic networks. We constrain the actor to be $\mathbb G$-equivariant and the critic to be $\mathbb G$-invariant, following the assertions made for MDPs in \Cref{subsec:background_symmetries}.
We do so by predefining the symmetric state and action representations $\rho_{\mathcal X}$ and $\rho_{\mathcal U}$. For example, in the robot's joint space, due to the $\mathbb C_2$ symmetry, a state $(q, \dot q)$ can be reflected to obtain the symmetric state $(g_s \triangleright q, g_s \triangleright \dot q)$ \cite{Ordonez2025}.
This actor-critic architecture is similar to that proposed in \cite{Su2024}, but we extend it by defining the symmetries for the privileged latent-state observations.

We also apply symmetry priors to the latent state.
The Koopman autoencoder learns to encode and decode the Koopman state using the symmetry information for the state $x_k$ and action $u_k$.
This is made possible by constraining the encoder $f_{\phi}$ and decoder $f_{\phi}^{-1}$ to be $\mathbb G$-equivariant. Using the notation from \Cref{subsec:background_symmetries}, the symmetry rules for the actor, critic, encoder, and decoder are summarized as: 
\begin{equation}
    \begin{aligned}
        \pi_{\theta}(g \triangleright x) &= g \triangleright \pi(x), \\
        V_{\xi}(g \triangleright x) &= V_{\xi}(x), \\
        f_{\phi}(g \triangleright x) &= g \triangleright z, \\
        f_{\phi}^{-1}(g \triangleright z) &= g \triangleright x.
    \end{aligned}
\end{equation}
Structurally, all these networks are termed \textit{equivariant multilayer perceptrons}, or EMLPs.
\subsection{Koopman Autoencoder with Symmetries}

To learn a lifting function, we train an equivariant autoencoder concurrently with the agent.
This simultaneous learning strategy avoids 
pretraining and
ensures training data are relevant to the state and action space regions explored by the agent since the control policy changes over the course of the training. 
We adapt the Equivariant Dynamics Auto-Encoder (eDAE) structure proposed in \cite{Apraez2024}, which is based on the original DAE \cite{Lusch2018}, to learn the dynamics of a controlled system, which we term an Equivariant Controlled Dynamics Auto-Encoder (ecDAE).

Thus, the linearized dynamics take the form of \Cref{eq:controlled_koopman}. 
We now represent the lifting function $\psi(x)$ as $f_{\phi}$ and the state reconstruction matrix $C$ as $f_{\phi}^{-1}$ to show that they are parameterized as neural networks in the autoencoder framework. 
We also employ a multi-step Koopman prediction, so we can adapt \Cref{eq:controlled_koopman} to reflect these specifics of our approach, as follows:
\vspace{-0.25cm}
\begin{equation}
    \begin{aligned}
        z_{k+H} &= A^{H} f_{\phi}(x_k) + \sum_{i=0}^{H-1} A^{H - 1 - i} B u_{k+i}, \\
        x_k &= f_{\phi}^{-1} (z_k).
    \end{aligned}
    \label{eq:ae_koopman_dyn}
\end{equation}

We use 
standard losses for training the autoencoders:
\begin{equation}
\begin{aligned}
        \mathcal L_{sr} &:= \| x_k - f_{\phi}^{-1}\left(f_{\phi}(x_k)\right)\|^2_2, \\
        \mathcal L_{lp} &:= \| f_{\phi}(x_{k+H}) - \mathcal K^{H} f_{\phi}(x_k)\|^2_2, \\
        \mathcal L_{sp} &:= \| x_{k+H} - f_{\phi}^{-1}\left(\mathcal K^{H} f_{\phi}(x_k)\right)\|^2_2.
    \end{aligned}
    \label{eq:ae_loss}
\end{equation}

\Cref{eq:ae_koopman_dyn} shows how the control input fits into the autoencoder structure, but does not address the symmetries.
$f_{\phi}(x) = z$ and $\pi_{\theta}(x) = u$ are constrained to be $\mathbb G$-equivariant by ensuring that the basis functions are $\mathbb G$-equivariant. Thus $\mathcal F(\mathcal X)$ becomes a symmetric function space, which can be decomposed into isotypic subspaces ${\mathcal F(\mathcal X) := \bigoplus^p_{i=1} \mathcal F_i(\mathcal X)}$, where $p$ is the number of distinct irreducible representations of $\mathbb G$ \cite{Ordonez2025}. This means that $A$ and $B$ are block-diagonal with $p$ blocks. Essentially, \Cref{eq:ae_loss} learns a Koopman operator $\mathcal K_i$ for each of these subspaces, and symmetry constraints are preserved throughout.

\subsection{Latent State Privileged Observations}
In standard actor-critic frameworks, the state $x$ is used as input to both the actor and the critic. 
Our actor-critic method is asymmetric (e.g., \cite{Pinto2018}), so the critic receives a single \emph{privileged} Koopman state prediction $z_{k+1} = A f_{\phi}(x_k) + B u_k$ learned by the ecDAE. Since the actor only requires $x_k$ as input, this architectural choice informs training without increasing computational overhead at deployment time.
The autoencoder is trained to predict the linear dynamics of the state $z$ over an $H$-step horizon. 
This property means it provides the critic with an informative predictive prior.

\subsection{Online DAE Training}
To avoid costly pretraining and to ensure that the explored state space is represented in the learned Koopman model, we train the ecDAE  concurrently with the RL policy.
At the beginning of training, rollouts of random actions are performed to fill a replay buffer with training data for the DAE. This buffer is distinct from the buffer used for training the actor and critic in PPO. At each RL training step, an ecDAE training step is also executed. We choose to store and select training data from a Prioritized Experience Replay (PER)~\cite{Schaul2015} buffer $\mathcal D(x_k, u_k, x_{k+1})$ for each iteration. 
PER prioritizes keeping buffer data which results in high loss.

\section{EXPERIMENTAL SETUP}

We now show that SKooP is modular and succeeds in improving performance, sample efficiency, and generalization to unseen scenarios. 
We present the results of training SKooP on multiple challenging bipedal locomotion tasks 
to empirically evaluate its capabilities.
In our experiments, we adopt vanilla PPO as an actor-critic algorithm due to its reliability and widespread use. However, note that SKooP is compatible with any actor-critic algorithm.
We test on multiple bipedal locomotion tasks inspired by~\cite{Li2024b}
for the Cyberdog~2 quadruped~\cite{Cyberdog2}. 
We consider 3 tasks, all of which require the quadruped to transition into an upright, bipedal position and walk on its hind legs: \emph{stand dance}, \emph{walk slope}, and \emph{push door}. 
Details on task parameters are available in \Cref{tab:task_changes} (Appendix). 
%
We employ the massively parallel Isaac Gym~\cite{Makoviychuk2021} physics simulator and execute our experiments on an NVIDIA A100 GPU.
Averaged over 100 iterations, a single training iteration of SKooP, PPO, and PPOeqic respectively for the \emph{stand dance} task takes 2.651~s, 1.978~s, and 2.245~s. PPOeqic takes 13.5\% longer than baseline PPO, and SKooP takes 34.0\% longer.



To apply our symmetry and Koopman techniques, we utilize the open source libraries DynamicsHarmonicsAnalysis~\cite{Apraez2024}, MorphoSymm~\cite{Ordonez2025}, and ESCNN~\cite{Cesa2022}. 
We employ a horizon of $H=5$ for our ecDAE training. We use a PER buffer with 10000 data points to train the autoencoder. 
Each task is trained with~8192 parallel environments, over~5 random seeds. The dimensionality ratio of the latent state $z$ to the state $x$ is~3. The \emph{stand dance} and \emph{walk slope} tasks are trained for 30000 iterations and have a state dimension of~47, while \emph{push door} is trained for 18000 iterations and has a state dimension of~43. All tasks use an action space of size 12, equal to the number of controllable joints. For further details on the task, refer to~\cite{Su2024}.



\section{RESULTS}

\begin{table*}[t]
\vspace{0.2cm}
\centering
\caption{Push door ablation study, showing (\% mean $\pm$ std).}
\vspace{-0.2cm}
\label{tab:si_ablation}
\begin{tabular*}{0.91\linewidth}{lllllll}
\toprule
\textbf{Method} & \textbf{Right SR} $\uparrow$ & \textbf{Left SR} $\uparrow$ & \textbf{SI} $\downarrow$ & \textbf{OOD Right SR} $\uparrow$ & \textbf{OOD Left SR} $\uparrow$ & \textbf{OOD SI} $\downarrow$ \\
\midrule
\textbf{PPO} & 70.01 ± 6.65 & 0.00 ± 0.00 & 199.99 ± 0.02 & 42.27 ± 7.65 & 0.01 ± 0.02 & 199.94 ± 0.12 \\
\textbf{PPOeqic} & 39.65 ± 5.07 & 39.12 ± 6.31 & 13.82 ± 6.88 & 21.43 ± 4.44 & 21.72 ± 4.10 & 17.68 ± 13.44 \\
\midrule
\textbf{SKooP-NoSym-NoPred} & \textbf{84.18} ± 4.31 & 0.02 ± 0.02 & 199.91 ± 0.10 & 45.95 ± 6.56 & 0.02 ± 0.02 & 199.82 ± 0.16 \\
\textbf{SKooP-NoSym} & 83.36 ± 3.86 & 0.05 ± 0.02 & 199.74 ± 0.09 & 41.23 ± 4.84 & 0.01 ± 0.01 & 199.89 ± 0.08 \\
\textbf{SKooP-NoPred} & 74.37 ± 12.88 & 68.45 ± 11.32 & \textbf{8.03} ± 6.89 & 44.94 ± 7.61 & 43.43 ± 5.90 & 9.14 ± 10.57 \\
\midrule
\textbf{SKooP} & 82.77 ± 2.96 & \textbf{77.70} ± 6.66 & 8.49 ± 4.26 & \textbf{47.65} ± 2.59 & \textbf{46.18} ± 4.37 & \textbf{9.06} ± 9.24 \\
\bottomrule
\end{tabular*}
\vspace{-0.4cm}
\end{table*}

\begin{figure*}
    \centering
    \includegraphics[width=\linewidth]{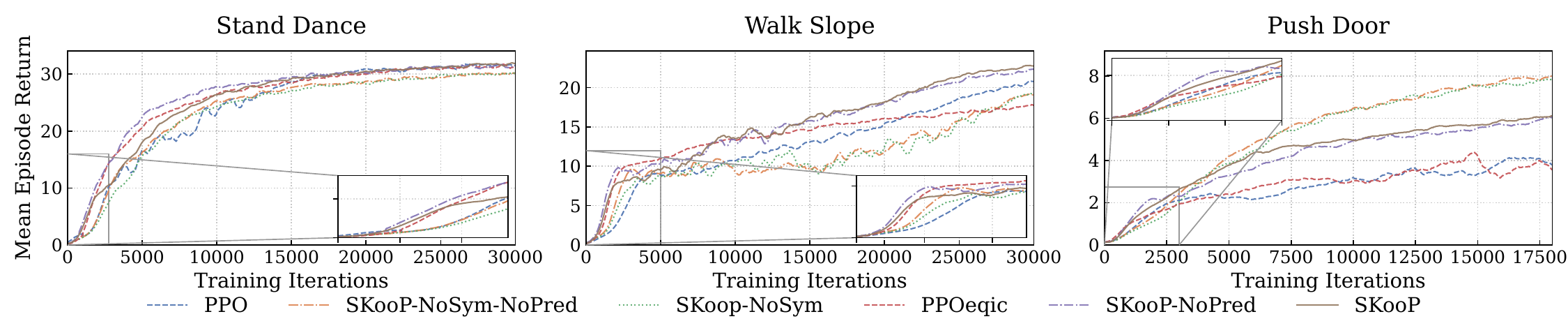} 
    \vspace{-0.75cm}
    \caption{Ablation 
    of training reward 
    applied to the following bipedal tasks: dancing, climbing a slope, and opening a door, each averaged over 5 seeds.} 
    \label{fig:reward_curves}
    \vspace{-0.5cm}
\end{figure*}



We define and adopt several standard metrics to evaluate our approach, including the mean episode return, training value loss, success rate (SR), symmetry index (SI), stability, and controllability.
We also evaluate SR and SI when starting in out-of-distribution (OOD) initial states.
Additionally, we provide an ablation study to isolate the effects of the symmetry constraints, latent state critic input, and Koopman prediction prior. Examples of SKooP policy performance on the considered tasks are available in the accompanying video.

The nomenclature we use for the ablation is as follows: \textit{PPO} refers to vanilla PPO, \textit{PPOeqic} denotes equivariant actor and invariant critic without using the autoencoder (presented in~\cite{Su2024}), and \textit{SKooP} denotes our complete approach. \textit{SKooP-NoSym-NoPred} is our approach without symmetries (using the controlled autoencoder, cDAE) and using $z_k$ instead of the prediction $z_{k+1}$, \textit{SKooP-NoPred} also uses $z_k$, but is with symmetries, while \textit{SKooP-NoSym} uses the prediction $z_{k+1}$ and no symmetries.

\subsection{Policy Symmetry Evaluation}

We select the \emph{push door} task as a means of evaluating the generalization of the policy learned by our method. 
To do this, we collect the SR, SI, OOD SR, and OOD SI metrics over 10000 episodes, averaged over 5 seeds per setting.
We define a successful episode as one in which the agent manages to open the door to at least $60^{\circ}$ and walk through it. 
The SR is the percentage of episodes in which the agent succeeds.
The SI is defined using the SR.
Specifically, $\text{SI} := \frac{2|X_R-X_L|}{X_R+X_L} \times 100\%$,
where $X_R$ is the SR for a scenario where the door opens on the right, while $X_L$ is the SR for the mirrored task which differs only in that the door opens on the left.
To evaluate OOD performance, the initial base orientation of the robot is sampled uniformly at random within $\pm 15^{\circ}$ around each axis compared to the trained initial orientation.
These metrics are similar to those used in~\cite{Su2024}. 
All models are trained only in the scenario with the right-opening door, while they are tested with both right- and left-opening doors to evaluate generalization to symmetric tasks.
The results are reported in \Cref{tab:si_ablation}.

\subsubsection{Success Rate (SR) and Symmetry Index (SI)}
\label{sec:push_door_si}
We find that all approaches that do not employ symmetry constraints, i.e., PPO, SKooP-NoSym-NoPred, and SKooP-NoSym, exhibit a very high SI, meaning that they do not generalize at all to the mirrored task (left-opening door). 
SKooP-NoSym and SKooP-NoSym-NoPred display the highest average SR for the right-side training task, although not significantly higher than SKooP.
However, they systematically fail on the left-side task.
Conversely, PPOeqic, which constrains the actor and critic with symmetry priors, achieves a diminished right-side SR compared to the baseline PPO, but generalizes well to the mirrored task. 
Both SKooP-NoPred and SKooP enforce symmetry priors on the actor and critic in the same way that PPOeqic does. SKooP-NoPred and SKooP additionally enforce the symmetry constraints in the ecDAE, leading to a latent space 
with embedded symmetries.
The results show that the use of symmetry priors in the privileged $z_k$ or $z_{k+1}$ input enables significantly higher SRs on both the right and left sides, and lower SI, with respect to PPOeqic. 
Compared to SKooP-NoPred, SKooP yields stronger performance across the board and similarly low SI, demonstrating the benefit of the Koopman prediction $z_{k+1}$. We note that the results for SKooP also exhibit lower variance compared to the other methods.
\Cref{fig:lr_push_door} visually compares the SKooP method on the original and mirrored door-opening tasks, showing similar behavior in both. 

\begin{figure*}
\vspace{0.25cm}
    \centering
    \includegraphics[width=0.95\linewidth]{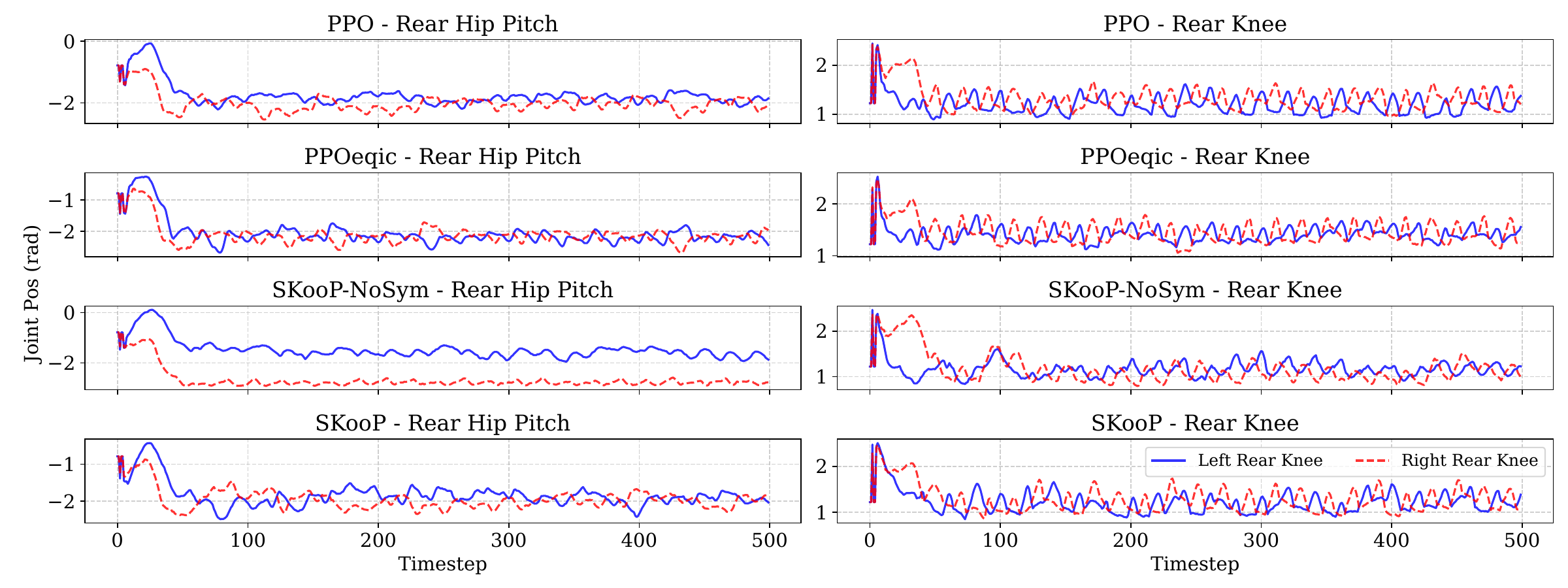}
    \vspace{-0.45cm}
    \caption{Comparison of \emph{stand dance} rear leg joint angles for PPO, PPOeqic, SKooP-NoSym, and SKooP.}
    \label{fig:stand_dance_joint_angles}
    \vspace{-0.4cm}
\end{figure*}

\begin{table}[t]
\centering
\caption{\emph{stand dance} steady-state gait metrics
averaged over 5 seeds.}
\vspace{-0.2cm}
\resizebox{\linewidth}{!}{
\begin{tabular}{llcccc}
\toprule
\textbf{Method} & \textbf{\makecell{Mean Phase\\ MAE (rad)}} & \textbf{Mean $\Delta \theta$ (rad)} & \textbf{Mean $\Delta \theta_{\text{diff}}$  (rad)} \\
\hline
\addlinespace
PPO & $0.216 \pm 0.006$ & $0.761 \pm 0.027$ & $0.100 \pm 0.042$ \\
PPOeqic & $0.136 \pm 0.012$ & $0.800 \pm 0.061$ & $0.103 \pm 0.000$ \\
SKooP-NoSym & $0.521 \pm 0.179$ & $0.854 \pm 0.222$ & $0.237 \pm 0.142$ \\
SKooP & $0.170 \pm 0.015$ & $0.813 \pm 0.049$ & $0.137 \pm 0.054$ \\
\bottomrule
\end{tabular}
}
\label{tab:gait_metrics}
\end{table}

\begin{table}[t]
\vspace{-0.1cm}
\centering
\caption{Summary of learned Koopman model stability and controllability metrics, averaged over 5 seeds.}
\vspace{-0.2cm}
\resizebox{\linewidth}{!}{
\begin{tabular}{l l c c c}
\toprule
\textbf{Task} & \makecell[l]{\textbf{DAE} \\ \textbf{Type}} & \textbf{$|\lambda(A)|_{max}$} & \textbf{$\kappa(\mathcal C)$} & \textbf{$\text{rank}(\mathcal C)$}/Total \\
\midrule
\textbf{Stand} & cDAE & 0.557 $\pm$ 0.012 & (1.61 $\pm$ 1.16)e+09 & (69.4 $\pm$ 2.9)/141 \\
\textbf{Dance} & ecDAE & 0.613 $\pm$ 0.019 & (1.99 $\pm$ 0.48)e+09 & (55.8 $\pm$ 1.8)/141 \\
 \midrule
\textbf{Walk} & cDAE & 0.551 $\pm$ 0.023 & (1.77 $\pm$ 0.89)e+09 & (68.6 $\pm$ 3.1)/141 \\
\textbf{Slope} & ecDAE & \textbf{1.013 $\pm$ 0.018} & (1.26 $\pm$ 2.41)e+11 & (45.2 $\pm$ 13.5)/141 \\
  \midrule
\textbf{Push} & cDAE & 0.572 $\pm$ 0.017 & (3.96 $\pm$ 0.70)e+08 & (71.0 $\pm$ 1.4)/129 \\
\textbf{Door} & ecDAE & \textbf{1.000 $\pm$ 0.003} & (8.17 $\pm$ 9.07)e+09 & (54.6 $\pm$ 1.5)/129 \\
\bottomrule
\end{tabular}
}
\vspace{-0.35cm}
\label{tab:a_eig_sv_summary}
\end{table}

The takeaways from these results are that, by providing a Koopman state prediction to the critic, and by constraining the actor, critic, and autoencoder with symmetry information, we enable the agent to learn not just a more successful policy, but also one that can generalize to a previously unseen, mirrored task.

\subsubsection{OOD Performance}

When tested on the OOD initial states that we define, we observe that while  SKooP-NoSym-NoPred and SKooP-NoSym  still perform comparably with PPO on the right-hand opening task, they also similarly fail on the mirrored task.
In contrast, PPOeqic, SKooP-NoPred, and SKooP consistently perform highly symmetrically even on OOD tasks, as revealed by the low SI values. 
Notably, in the OOD case SKooP significantly outperforms PPOeqic for both SR and SI, also achieving stronger mean performance and lower variance with respect to SKoop-NoPred.
\subsubsection{Joint Trajectory Analysis}

We also examine a sample of joint trajectories for the rear legs in  the \emph{stand dance} task (see \Cref{fig:stand_dance_joint_angles}). 
Symmetric motion of the rear legs indicates greater stability of the robot.
In the plots, the motion is symmetric if the right and left joint trajectories oscillate in similar ranges.
The results show that SKooP produces the most symmetric behavior.
Instead, PPO learns to command larger angles for the right rear knee than for the left.

\Cref{tab:gait_metrics} reports the phase mean absolute error (MAE), the mean step amplitude $\Delta \theta$, and the amplitude difference between left and right joints $\Delta \theta_{\text{diff}}$, for the considered  methods.
The phase MAE is defined as $\frac{1}{T} \sum_{k=0}^{T} |q_L(k) - q_R(k+ \tau)|$,
where $T$ is the total number of timesteps, $q_L, q_R$ are the left and right joint positions of a given joint, and $\tau=1/(2f_d \Delta t)$ is the expected phase offset assuming that the ideal is a
$180^{\circ}$ offset, given the desired gait frequency $f_d=2.5 \si{\hertz}$ and timestep $\Delta t$. SKooP-NoSym learns significantly different ranges for the left and right rear hip pitch joints, which is expected since it does not have access to symmetry priors. \Cref{tab:gait_metrics} shows that SKooP rectifies this with the use of symmetry priors, reducing the mean phase MAE by 96.7\% and $\Delta \theta_{\text{diff}}$ by 42.3\% in comparison.
While PPOeqic learns more visually symmetric right and left joint ranges, the left rear knee consistently has a larger maximum angle than the right one.
Furthermore, SKooP learns joint trajectories that oscillate within the same range for both the left and right~joints. Finally, \Cref{tab:gait_metrics} also shows that SKooP achieves larger and more consistent $\Delta \theta$ across seeds than PPOeqic.

\subsection{Training Performance} 
Across the three tasks, \Cref{fig:reward_curves} shows that
the SKooP, SKooP-NoPred, and PPOeqic approaches incorporating symmetries have steeper initial convergence rates than competitors.
We note that across the tasks, the -NoPred methods which learn with $z_k$ converge slightly faster than their counterparts which learn with the Koopman prediction $z_{k+1}$. 
This is explained by the fact that during the initial learning phase the autoencoder is exposed to a wide variety of states and actions and has not yet converged.
Using only the lifting function to get $z_k$ introduces less error in this phase than using both the lifting function and the  Koopman model prediction to get $z_{k+1}$. 
For the \emph{walk slope} task, incorporating the latent state (predicted or not) is sufficient to improve the convergence rate without the use of symmetries, while symmetries provide further improvement.

Overall, SKooP achieves 
higher returns than PPO and 
PPOeqic.
The easiest task to learn, \emph{stand dance}, exhibits comparably small improvements from the use of Koopman prediction and symmetry priors. Instead, on the most challenging, asymmetric task, \emph{push door}, PPOeqic provides faster initial convergence than PPO while achieving similar reward.
Interestingly, for \emph{push door}, SKooP-NoSym and SKooP-NoSym-NoPred learn to maximize rewards and become highly specialized. However, as discussed in~\Cref{sec:push_door_si}, they fail to generalize to the mirrored task. On \emph{push door}, SKooP significantly outperforms PPO and PPOeqic in terms of reward values and converges faster than SKooP-NoPred.

\subsection{Stability and Controllability of the Koopman Model}


A discrete-time linear system with eigenvalues $\lambda(A)$ closer to the center of the unit circle has dynamics that quickly decay to an equilibrium point. Conversely, at the edge of the unit circle, eigenvalues are marginally stable, meaning the system has dynamics that are time invariant. 

\begin{figure}
\vspace{0.2cm}
    \centering
    \includegraphics[width=0.9\linewidth]{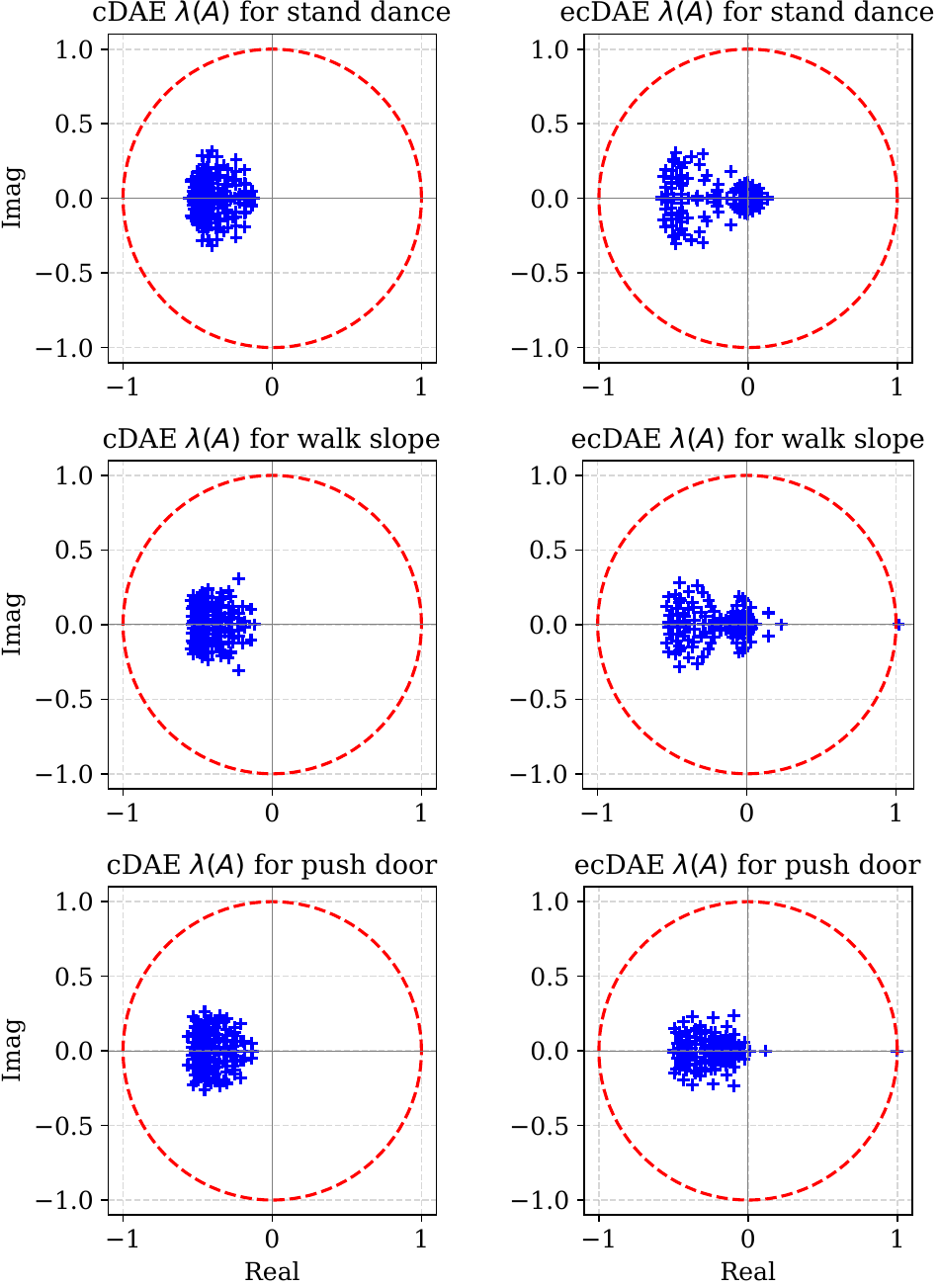}
    \vspace{-0.3cm}
    \caption{Eigenvalues of selected learned Koopman state matrices $A$.}
    \label{fig:eigval_study}
    \vspace{-0.2cm}
\end{figure}

Remarkably, \Cref{tab:a_eig_sv_summary} and \Cref{fig:eigval_study} show that the use of symmetry constraints always produces a larger $|\lambda(A)|_{max}$, with low variance across seeds. 
In fact, in two out of three tasks, the ecDAE learns exactly one eigenvalue $|\lambda(A)| \approx 1.0$. 
\Cref{tab:a_eig_sv_summary} further reports numerical values related to the controllability, which measures how well the controller can reach goal states in the state space of the Koopman model. 
This can be evaluated using the singular values $\sigma$ of the controllability matrix $\mathcal C = \begin{bmatrix} B & AB & \cdots & A^{n-1} B \end{bmatrix}$.
Across all tasks and seeds, the ecDAE has a larger condition number $\kappa(\mathcal C)=\sigma_{max}(\mathcal C)/\sigma_{min}(\mathcal C)$ and lower controllability rank $\text{rank}(\mathcal C)$ than its cDAE counterpart. This suggests that the ecDAE learns a more compact representation, by recognizing that symmetric modes cannot be controlled independently. The larger condition number shows that the ecDAE is forcing redundant control directions towards 0.

\begin{table}[t]
\centering
\caption{MSE comparison of cDAE and the symmetry-constrained ecDAE for \emph{push door}, with ablation of invariant mode, averaged over 5 seeds.}
\vspace{-0.2cm}
\resizebox{0.95\linewidth}{!}{
\begin{tabular}{lcc}
\toprule
\textbf{Metric} & \textbf{cDAE} & \textbf{ecDAE}\\
\midrule
\textbf{$\mathcal L_{sr}$ MSE} & $1.503 \pm 0.153$ & \textbf{$1.245 \pm 0.152$} \\
\textbf{1-Step $\mathcal L_{lp}$ Latent MSE}  & \textbf{$0.283 \pm 0.014$} & $0.274 \pm 0.017$ \\
\textbf{5-Step $\mathcal L_{sp}$ MSE (Right)} & $2.144 \pm 0.078$ & \textbf{$1.778 \pm 0.153$} \\
\textbf{5-Step $\mathcal L_{sp}$ MSE (Left)}  & $3.150 \pm 0.143$ & \textbf{$1.848 \pm 0.196$} \\
\textbf{Invariant $\lambda$} & N/A & \textbf{$1.000 \pm 0.003$} \\
\midrule
\multicolumn{3}{c}{\textbf{Ablation}}\\
\midrule
\textbf{Ablated $\mathcal L_{sp}$ MSE (Right)}& N/A & $3.054 \pm 1.047$ \\
\textbf{Ablated $\mathcal L_{sp}$ MSE (Left)} & N/A & $3.524 \pm 1.403$ \\
\textbf{Only Inv $|\lambda|$ $\mathcal L_{sp}$ MSE (Right)}& N/A & $3.186 \pm 1.265$ \\
\textbf{Only Inv $|\lambda|$ $\mathcal L_{sp}$ MSE (Left)} & N/A & $3.555 \pm 1.549$ \\

\bottomrule
\end{tabular}
}
\label{tab:dae_comparison}
\vspace{-0.5cm}
\end{table}
In order to better understand the performance and behavior of the cDAE vs. the ecDAE, \Cref{tab:dae_comparison} reports the average autoencoder state reconstruction, 1-step latent prediction, and $H$-step state prediction errors for a horizon $H=5$ (\Cref{eq:ae_loss}) for cDAE and ecDAE trained for \emph{push door}. The table also shows the value of the invariant mode (if any) and compares the state prediction error on the trained (right) and mirrored task (left) both with the invariant mode enabled, disabled, and with only the invariant mode enabled. This analysis shows that while the symmetry constraints may have no significant benefit for latent state prediction, they improve state reconstruction and state prediction. The use of symmetry priors also enables the ecDAE to achieve almost identical state prediction on the mirrored task as it does on the trained task, whereas the cDAE's state prediction is around 50\% worse on the mirrored task. Furthermore, removing the ecDAE's invariant mode significantly increases $\mathcal L_{sp}$ MSE, as does disabling all the other modes except for the invariant one.
Since only the ecDAE learns this $|\lambda| \approx 1$, it must help represent the invariance of the system dynamics across symmetry groups. Unlike the other tasks, \emph{stand dance} does not require significant base displacement, which may explain why for that task the ecDAE does not identify an invariant mode to represent the symmetry.

\subsection{Sim-to-sim}

To validate the robustness of the learned SKooP policies, we evaluate sim-to-sim transfer from the training simulator, Isaac Gym, to Mujoco. Unlike Isaac Gym, which trades some physical accuracy for massively parallelized training efficiency, Mujoco's high-fidelity solver better models real-world joint and contact physics. \Cref{tab:sim2sim_robustness} compares the command tracking MSE for the \emph{stand dance} task across both simulators. Because this task is governed by linear $x, y$ and angular $z$ velocity commands, we report the linear $v_{xy}$ and angular $\omega_z$ tracking errors.
To test robustness, we evaluate conditions at the edge of the agent's training distribution. The training ranges are: friction coefficient $[1.0, 3.0]$, added mass $[-0.5, 0.5]$~\si{\kilogram}, velocity pushes $[0.0, 0.2]$ \si{m/s}, and rotor inertia $0.0$ \si{\kilogram \meter \squared}. Unless stated otherwise, the default test settings are: friction 1.0, added mass $0.0$ \si{\kilogram}, velocity push $0.0$ \si{m/s}, and rotor inertia $0.005$ \si{\kilogram \meter \squared}. Linear velocity commands $v_{xy}$ are set to $0.0$. We note that Isaac Gym by default assigns the rotor inertia (armature) to $0.0$ \si{\kilogram \meter \squared} during training, and in Mujoco this causes failure. 

None of the rollouts summarized in  \Cref{tab:sim2sim_robustness} result in failure. However, they show that while the policy produces similar linear velocity MSE across test conditions in both simulators, it suffers in the yaw velocity tracking. These results demonstrate that SKooP is robust to perturbations and transferable across simulated settings.


\begin{table}[t]
\vspace{0.25cm}
\centering
\caption{Study of SKooP Sim-to-Sim Robustness on \emph{stand dance} Linear $v_{xy}$ and Angular $\omega_z$ Command Tracking Performance (MSE).}
\vspace{-0.2cm}
\label{tab:sim2sim_robustness}
\addtolength{\tabcolsep}{-2pt}
\resizebox{0.9\linewidth}{!}{
\begin{tabular}{l c c c c}
\toprule
\textbf{Test Condition} & \multicolumn{2}{c}{\textbf{$v_{xy}$ MSE}} & \multicolumn{2}{c}{\textbf{$\omega_z$ MSE}} \\
& Isaac Gym & MuJoCo & Isaac Gym & MuJoCo \\
\midrule
Friction $\mu=0.4$ & 0.0341 & 0.0245 & 0.2795 & 0.6017 \\
Friction $\mu=1.0$ & 0.0271 & 0.0231 & 0.1719 & 0.5048 \\
Mass Offset 0.5 \si{kg} & 0.0273 & 0.0257 & 0.1656 & 0.2904 \\
Push Vel 0.3 \si{m/s} & 0.0320 & 0.0257 & 0.1981 & 0.5325 \\
Armature 0.002 & 0.0308 & 0.0229 & 0.1494 & 0.4826 \\
Armature 0.005 & 0.0406 & 0.0227 & 0.2339 & 0.4986 \\
\bottomrule
\end{tabular}
}
\vspace{-0.6cm}
\end{table}

\section{CONCLUSIONS}
We introduce SKooP, a unified learning approach which accelerates convergence and improves reward performance. 
Imposing symmetry constraints enables SKooP to generalize beyond training scenarios,
while ecDAE-trained Koopman predictions for value function learning improves total return and qualitative learned behavior.
We also demonstrate that SKooP is modular and produces consistent advantages across tasks.
Future work will address alternative Koopman learning strategies and transfer to other robots and tasks.

\section*{Appendix}
\label{sec:appendix}

Here, we include a table listing some of the parameter settings we use for the quadruped training tasks we consider.

\begin{table}[h]
\centering
\caption{Summary of task settings.
Remaining parameters as in~\cite{Su2024}.}
\vspace{-0.2cm}
\label{tab:task_changes}
\begin{tabular*}{0.82\linewidth}{lllllll}
\toprule
\textbf{Task} & \textbf{Parameter}& \textbf{Value} \\
\midrule
\textbf{Stand dance}  & Lin. vel. tracking weight & 0.8 \\
\textbf{Stand dance}  & Ang. vel. tracking weight & 0.5 \\
\textbf{Stand dance}  & Upright weight & 1.2 \\
\textbf{Stand dance}  & Lift up weight & 0.8 \\
\midrule
\textbf{Push door}    & Lin. vel. $x$ command range & [0.15, 0.3] \\
\bottomrule
\end{tabular*}
\end{table}

\addtolength{\textheight}{-0cm}   



\vspace{-0.5cm}

\bibliographystyle{IEEEtran}
\bibliography{IEEEabrv,symm_koopman_critic}

\end{document}